\documentclass[11pt]{article}

\usepackage[margin=1in]{geometry}
\usepackage{times}
\usepackage{amsmath,amssymb}
\usepackage{booktabs}
\usepackage{graphicx}
\usepackage{hyperref}
\usepackage[round]{natbib}
\usepackage[font=small,labelfont=bf,labelsep=period]{caption}
\usepackage{xcolor}
\usepackage{microtype}

\hypersetup{
  colorlinks=true,
  linkcolor=black,
  citecolor=black,
  urlcolor=black
}

\newcommand{\R}{\mathbb{R}}

\title{\textbf{Five Architectures, One Shared Space} \\[0.3em] \large Cross-Model Representational Alignment Without Weight Merging}

\author{
  Anthonio Oladimeji Gabriel$^{1}$ \quad
  Dimeji Olawuyi$^{2,3}$ \quad
  Toba Ajayi$^{2,3}$ \quad
  Temilola Aderemi$^{2,3}$ \\[0.6em]
  {\small $^{1}$Centre for Clinical Intelligence and Safety, \`{I}y\`{a}w\'{o}} \\
  {\small $^{2}$\`{I}y\`{a}w\'{o}} \\
  {\small $^{3}$Faculty of Clinical Sciences, College of Medicine, University of Ibadan} \\[0.4em]
  {\small Correspondence: anthoniooladimeji11@gmail.com}
}
\date{October 2026}

\begin{document}
\maketitle

\begin{abstract}
\noindent Combining language models is usually done by averaging weights, which works only when the models share a common starting point. We ask whether five frozen, independently trained models with different internal designs (four transformers from four organizations and one state-space model) can be related to one another through their activations alone, with no weight merging. Using permutation tests, an independent replication probe set, depth sweeps, pooling ablations and confound checks, we find statistically robust correspondence between how the five models organize the same text. It is strongest at each model's final layer and weaker for the state-space model, whose weakest pairing is still $35$--$43\times$ chance. We use this to build a shared space with Generalized Procrustes Analysis, in which no model is a fixed reference. At the final layer its mean held-out retrieval accuracy over the twenty ordered model pairs is $0.91$ to $0.93$ on two probe sets (chance $0.03$), beating pairwise linear maps ($0.82$ to $0.85$), but it is worse than those maps in the middle layers. The space transfers to ordinary English content of unfamiliar grammatical form and fails on mathematical and logical notation. We then test whether the space supports a router that picks among the five models on gradeable tasks. Across four task types we find no case where routing beats always using the strongest model, and we report this as a negative result. A correctness signal that looked strong turned out to be category membership in disguise. All experiments ran on one laptop without a GPU.
\end{abstract}

\section{Introduction}

The standard way to combine neural networks is to average their weights \citep{wortsman2022soups}. This works well, but it depends on the models sharing a weight space, typically because they are fine-tuned from one base checkpoint. Even for independently trained networks of identical architecture, merging requires first undoing permutation symmetries of the hidden units \citep{ainsworth2023gitrebasin}. None of this applies to models that differ in architecture, training data and parameter count.

We ask a different question. Rather than merging weights, can very different models be related through how they internally represent the same input, while every model stays frozen? If so, a small trainable component could in principle combine their outputs without touching their weights, preserving each model's original behavior exactly.

The question sits next to a growing body of evidence that independently trained networks tend toward similar representations as they scale \citep{huh2024platonic}, and next to the tools used to measure that similarity \citep{kornblith2019similarity,bansal2021stitching}. We study a case that is harder than most: five small open models trained by different groups, one of which (Mamba, \citealp{gu2023mamba}) is a state-space model that shares no sequence-mixing mechanism with the other four.

\paragraph{Contributions.}
\begin{enumerate}
  \item We show, using a series of experiments designed to falsify our own early results, that cross-model correspondence exists among these five models, and we state where it is strong, where it is weak, and which apparent signals were artifacts (Sections~\ref{sec:stage1}--\ref{sec:stage2}).
  \item We build a shared space with Generalized Procrustes Analysis (GPA), compare it with pairwise linear maps at every depth, and measure its generalization boundary (Sections~\ref{sec:construction}--\ref{sec:stage3}).
  \item We test whether the space has practical value for routing between the five models and find none, after a four-task search that included one task chosen because the literature predicted an architectural difference (Section~\ref{sec:stage4}).
\end{enumerate}
Everything was run on a single consumer laptop (Apple M1 Pro, 16\,GB, no discrete GPU). This limits scale, which we discuss in Section~\ref{sec:discussion}, but it makes every number here cheap to reproduce.

\begin{figure}[!htbp]
\centering
\includegraphics[width=\textwidth]{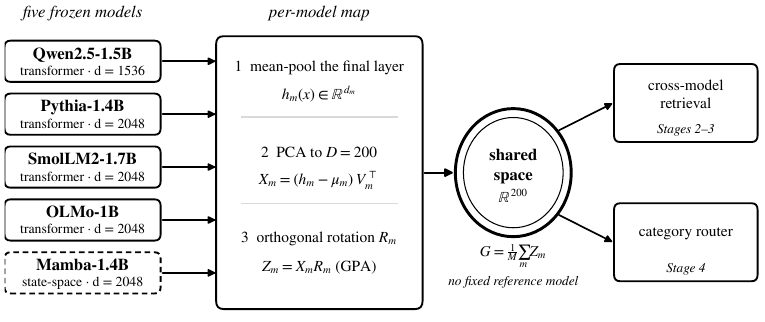}
\caption{Overview. Each frozen model's final-layer, mean-pooled activation is reduced by its own PCA, then rotated by its own orthogonal matrix into a space defined by the consensus of all five models. No model is the reference. The space is evaluated by retrieval (Stages 2 and 3) and used to build a category router (Stage 4). The dashed box marks the state-space model.}
\label{fig:overview}
\end{figure}

\section{Related Work}

\paragraph{Model merging.} Averaging the weights of fine-tuned checkpoints of one base model improves accuracy and robustness \citep{wortsman2022soups}. Merging independently trained networks of the same architecture can work after aligning hidden units up to permutation \citep{ainsworth2023gitrebasin}. Both assume a shared parameterization. Our models have different widths, depths and sequence-mixing mechanisms, so there is no parameter-level correspondence to exploit, and we work on activations instead.

\paragraph{Representational similarity.} Centered kernel alignment \citep{kornblith2019similarity} and model stitching \citep{bansal2021stitching} compare representations of different networks, with the latter asking whether a learned connector can join the early part of one network to the late part of another. The Platonic Representation Hypothesis \citep{huh2024platonic} argues that representations of independently trained models converge as they scale. Our setting is a small-scale test of that idea with one deliberately non-transformer model, and with the added question of whether the correspondence can be turned into an explicit common space.

\paragraph{Aligning separate spaces.} Orthogonal Procrustes alignment \citep{schonemann1966generalized} maps one point configuration onto another using only a rotation (and possibly a reflection), and has been used to align word-embedding spaces across languages \citep{conneau2018word}. Generalized Procrustes Analysis \citep{gower1975gpa} extends this to many configurations at once, aligning all of them to their own mean. Relative representations \citep{moschella2023relative} describe each input by its similarity to a set of anchor inputs, which makes latent spaces of different models directly comparable. We use GPA because it keeps every model on equal footing and uses only orthogonal maps, the most conservative family of transforms.

\paragraph{State-space models versus attention.} Selective state-space models such as Mamba \citep{gu2023mamba} match transformers on many language tasks at lower cost, but associative recall, retrieving the value paired with a key from context, is a documented weakness, because attention can index any position while a state-space model must compress history into a fixed-size state \citep{arora2023zoology}. We use this to design one targeted task in Section~\ref{sec:stage4}.

\paragraph{Routing.} Systems such as RouteLLM \citep{ong2024routellm} learn to choose between a strong and a weak model per query to save cost. We ask a different question: whether a space built without any task labels, from activations only, carries enough information to route among five models.

\section{Models and Experimental Setup}

\begin{table}[!htbp]
\centering
\caption{The five frozen models used throughout.}
\label{tab:models}
\begin{tabular}{lllrr}
\toprule
Model & Builder & Architecture & Layers & Hidden dim \\
\midrule
Qwen2.5-1.5B \citep{yang2024qwen25} & Alibaba & Transformer & 29 & 1536 \\
Pythia-1.4B \citep{biderman2023pythia} & EleutherAI & Transformer & 25 & 2048 \\
SmolLM2-1.7B \citep{allal2025smollm2} & Hugging Face & Transformer & 25 & 2048 \\
OLMo-1B \citep{groeneveld2024olmo} & AI2 & Transformer & 17 & 2048 \\
Mamba-1.4B \citep{gu2023mamba} & State-Spaces & State-space model & 49 & 2048 \\
\bottomrule
\end{tabular}
\end{table}

All five are base (not instruction-tuned) checkpoints, run in inference mode on one laptop. Mamba ran without accelerated kernels in our software environment and was about $25\times$ slower per probe than the transformers.

\paragraph{Probes.} We wrote 200 short inputs, 20 in each of ten categories: declarative facts, questions, sentence completions, syntactically ambiguous sentences, grammatical nonsense, logical reasoning, emotionally valenced text, source code, non-English text (six languages), and deliberate token repetition. A second 200-probe set (\emph{set 2}) with the same categories and entirely different content was written later and is used as an independent replication. We refer to the first as \emph{set 1}. Both sets were written by the authors, which we return to in Section~\ref{sec:discussion}.

\paragraph{Representation.} For model $m$ with $L_m$ layers and input $x$ of $T$ tokens, let $h^{(\ell)}_{m,t}(x)\in\R^{d_m}$ be the layer-$\ell$ hidden state at token $t$. The representation used throughout is the token-mean
\begin{equation}
h^{(\ell)}_m(x) = \frac{1}{T}\sum_{t=1}^{T} h^{(\ell)}_{m,t}(x).
\end{equation}
Unless stated otherwise $\ell=L_m$ (the final layer). Depth comparisons use relative depths $0,25,50,75,100\%$ of each model's own layer count. Mean pooling was chosen after comparing it with first-token, middle-token, last-token and maximum-norm-token pooling at each model's middle layer. For Qwen, Pythia and SmolLM the first-token and maximum-norm-token vectors were identical for every probe (an attention-sink effect), single-token schemes collapsed onto one axis for at least one scheme in each transformer, and mean pooling showed none of these artifacts. Mamba is the exception: its last-token vector separated the categories better than its mean-pooled one. We kept mean pooling as a uniform default and note that it may understate Mamba.

\paragraph{Agreement measures.} Let $D_m\in\R^{n\times n}$ be the matrix of pairwise cosine distances between the $n$ probe representations of model $m$. \emph{Distance-matrix agreement} between models $a$ and $b$ is the Spearman correlation $\rho_{ab}$ between the upper-triangular entries of $D_a$ and $D_b$. \emph{Nearest-neighbor match} is the fraction of probes $i$ for which $\arg\min_{j\ne i}D_a[i,j]=\arg\min_{j\ne i}D_b[i,j]$, i.e.\ both models name the same other probe as most similar; chance is $1/(n-1)$ ($0.0063$ for $n=160$). All nulls come from $2{,}000$ permutations of probe identity within one model; the permutation $p$-values are therefore bounded below by $5\times10^{-4}$. Unless noted, ``whole set'' analyses use the 160 probes of the eight categories other than code and multilingual text, which behave as outliers (Section~\ref{sec:stage2}).

\section{Stage 1: Per-Model Geometry}
\label{sec:stage1}

We projected each model's layer-wise activations to two dimensions with PCA. The four transformers each show a compressed plateau across their middle layers, where $98$--$99.8\%$ of variance lies in the first two components (Qwen: layers 5--20). Mamba shows no plateau: two-dimensional variance stays at $10$--$20\%$ through the middle of the network and rises to $56\%$ only at the final layer. All five models reorganize sharply at their own final layer, which is the layer where each prepares its output. In the middle layers even the four transformers look unlike one another, so aligning layer $k$ of one model with layer $k$ of another is not meaningful. This motivated analyses that treat depth as a variable (Section~\ref{sec:stage2}).

\section{Stage 2: Is There Cross-Model Correspondence?}
\label{sec:stage2}

\subsection{A first result that did not hold up}

We first asked whether the models agree on which \emph{categories} are similar. Using category centroids, the mean pairwise agreement over all ten categories was $\rho=0.914$. This number was inflated by code and multilingual text, which every model places far from everything else. On the remaining eight categories the mean agreement was $\rho=0.808$, and a permutation test that shuffled category labels within each model gave a null mean of $0.627$ ($\mathrm{sd}\ 0.117$), so that the observed value beat only $95.2\%$ of null draws ($p\approx0.048$). Random partitions of the same probes already agree strongly because of shared shallow statistics such as word length and word rarity. The category-level evidence was therefore marginal, and we did not build on it.

\subsection{Probe-level agreement}

\begin{figure}[!htbp]
\centering
\includegraphics[scale=0.88]{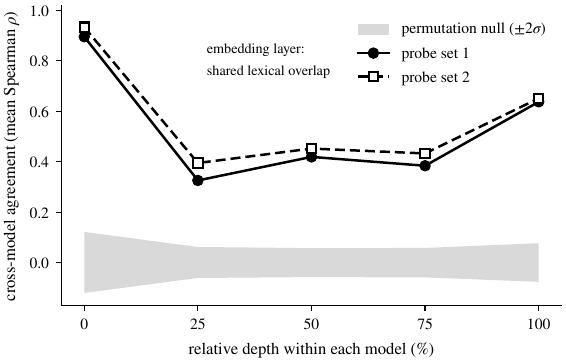}
\caption{Cross-model agreement by depth. Each point is the mean Spearman $\rho$ between distance matrices over all ten model pairs, on the 160 prose probes of each set. The grey band is $\pm2$ standard deviations of the permutation null. Agreement is high at the embedding layer for trivial lexical reasons, lowest at 25\% depth, and rises again at the final layer.}
\label{fig:depth}
\end{figure}

Working at the level of individual probes removes the averaging problem. At the final layer the mean agreement $\rho$ over the ten model pairs was $0.64$ (set 1) and $0.65$ (set 2), against permutation nulls centered on zero with standard deviation $0.03$ to $0.04$. For nearest-neighbor match, the ten-pair mean was $0.324$ on set 1 and $0.407$ on set 2, against a null of $0.0063$, which is $50\times$ and $65\times$ chance ($p<5\times10^{-4}$ for both). The result replicates on the second probe set (Figure~\ref{fig:depth}), is significant at every depth tested, and is not explained by probe length or lexical rarity (the correlation between each model's distance matrix and the length or rarity distance matrix was $\le0.10$ for four of the five models and at most $0.18$ for OLMo).

Two cautions apply. Agreement at relative depth $0\%$ (the embedding layer) is high ($0.90$ and $0.93$) but is expected from shared lexical overlap in any pair of models, and we do not treat it as evidence of shared processing. And because the probe-level null is tight, a very small $p$-value does not mean that the effect is equally strong at every depth: the effect size falls to $0.33$--$0.45$ in the middle of the network and recovers to $0.64$--$0.65$ at the final layer.

\subsection{The state-space model is weaker, in degree and not in kind}

\begin{figure}[!htbp]
\centering
\includegraphics[width=\textwidth]{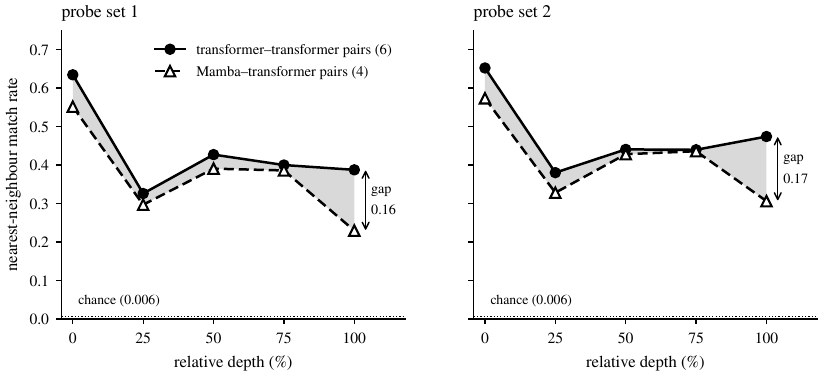}
\caption{Nearest-neighbor match rate by depth for transformer-only pairs and Mamba-involving pairs, on the two probe sets. The shaded gap is smallest in the middle of the network and largest at the final layer. Chance is $0.0063$.}
\label{fig:mamba}
\end{figure}

At the final layer, the four Mamba-involving pairs were the four weakest of the ten pairs in both probe sets (mean nearest-neighbor match $0.230$ against $0.388$ for the six transformer-only pairs on set 1, and $0.306$ against $0.474$ on set 2), with no overlap between the two groups. The weakest Mamba pair, Mamba with OLMo, still sits at $35$--$43\times$ chance. The size of the gap depends on depth (Figure~\ref{fig:mamba}): it is near zero at $75\%$ depth ($0.014$ and $0.004$) and largest at the final layer ($0.16$ and $0.17$), which is where the all-pairs agreement is strongest.

To see why the gap widens at the final layer we measured each model's own nearest-neighbor self-consistency between consecutive late layers. Pythia, SmolLM2 and OLMo each reorganize substantially on entering their last layer (self-consistency drops by $0.31$--$0.59$), while Mamba reorganizes mildly ($0.09$--$0.11$) and Qwen reorganizes least. The category-level changes made by the three reorganizing transformers are positively correlated with one another in both probe sets (Spearman $0.33$ to $0.77$ across the three pairs), whereas Mamba's changes correlate weakly or negatively with every other model ($-0.36$ to $0.34$), and Qwen's correlate inconsistently across sets. So the transformers that reorganize at the end do so in a shared direction, and Mamba does not move that way. We do not claim to know why. The explanation is descriptive, not causal.

\section{Constructing a Shared Space}
\label{sec:construction}

\subsection{Pairwise linear maps}

As a first construction we fit ordinary least-squares maps between every ordered pair of models (20 maps) and scored them with 5-fold cross-validated retrieval: given a held-out probe's mapped vector from model $a$, is the nearest neighbor (cosine) among the held-out probes of model $b$ the same probe? Each held-out fold has 32 probes, so chance is $1/32=0.031$. Accuracy at the final layer ranged from $0.70$ to $0.97$ over the 20 maps, with means of $0.816$ (set 1) and $0.854$ (set 2). Notably, Mamba pairs were not the weakest here (the single best map, at $0.97$, was between Mamba and SmolLM2), which suggests Mamba's activations are rotated or scaled differently from the transformers' and not incompatible with them, even though the raw nearest-neighbor comparison of the previous section ranked them lowest. Fitting every model onto one fixed target makes that target a privileged reference, and each map has up to $2048\times2048$ free parameters estimated from 128 training probes.

\subsection{Generalized Procrustes Analysis}

Let $X_m\in\R^{n\times D}$ be model $m$'s centered activations reduced by PCA to a common dimension $D$ (PCA is needed because $d_m$ differs between models):
\begin{equation}
X_m = \bigl(H_m-\mathbf{1}\mu_m^{\top}\bigr)V_m, \qquad V_m\in\R^{d_m\times D},\ V_m^{\top}V_m=I,
\end{equation}
where $H_m$ stacks the representations $h_m(x)$ and $\mu_m$ is their mean. GPA \citep{gower1975gpa} finds one orthogonal matrix per model, and a consensus configuration $G$, by solving
\begin{equation}
\min_{\{R_m\},\,G}\ \sum_{m=1}^{M}\bigl\lVert X_m R_m - G\bigr\rVert_F^2
\quad\text{s.t. } R_m^{\top}R_m=I,\qquad G=\frac{1}{M}\sum_{m=1}^{M}X_m R_m .
\end{equation}
It is solved by alternating two steps until the change in the objective is below $10^{-8}$ (maximum 50 iterations): given $G$, each $R_m$ is the orthogonal Procrustes solution \citep{schonemann1966generalized}, $R_m=U_mW_m^{\top}$ with $X_m^{\top}G=U_m\Sigma_m W_m^{\top}$ (reflections allowed, no scaling); given the $R_m$, $G$ is updated to their mean. The iteration starts from the first model's centered configuration, which is only an initialization, since every model is rotated toward the evolving mean. No model is the target; the space is the consensus. A new input from model $m$ enters the space as $z_m(x)=\bigl(h_m(x)-\mu_m\bigr)V_mR_m\in\R^{D}$.

\paragraph{Evaluation.} For held-out probes, the PCA basis $V_m$ and rotations $R_m$ are fit on the four training folds only (a first version fit PCA on all probes, which inflated accuracy by $0.007$ to $0.019$; the figures below use the corrected protocol). For each ordered pair $(a,b)$, accuracy is the fraction of held-out probes $i$ whose $z_a(x_i)$ has cosine-nearest neighbor $z_b(x_i)$ among the held-out $\{z_b(x_j)\}$.

\begin{figure}[!htbp]
\centering
\includegraphics[width=\textwidth]{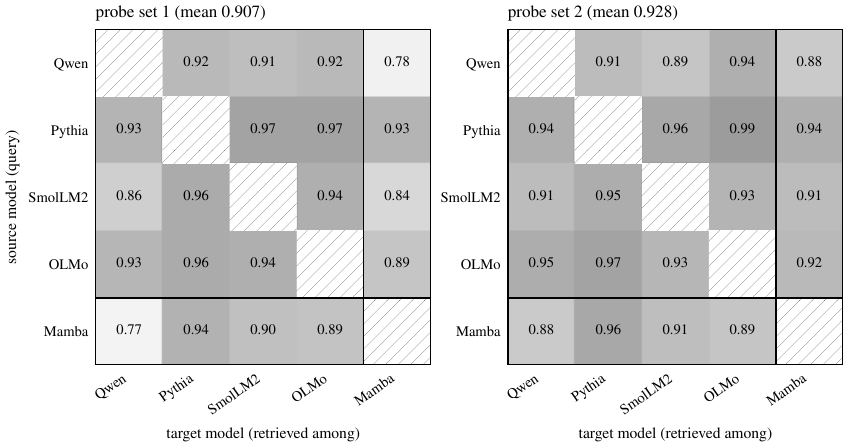}
\caption{Held-out retrieval accuracy for every ordered pair, through the GPA space (final layer, $D=100$, 5-fold cross-validation, chance $0.031$). Rows are the query model and columns the model retrieved among. Thick lines separate Mamba. The weakest cells involve Mamba and Qwen on set 1.}
\label{fig:heat}
\end{figure}

\paragraph{Results at the final layer.} With $D=100$, mean accuracy over the 20 ordered pairs was $0.907$ (set 1) and $0.928$ (set 2), against $0.816$ and $0.854$ for pairwise maps. Figure~\ref{fig:heat} shows every pair. All twenty ordered accuracies lie between $0.77$ and $0.99$, the two weakest (both under $0.80$ on set 1) being Qwen$\leftrightarrow$Mamba. Asymmetry between the two directions of a pair is small ($\le0.06$) compared with pairwise maps ($\le0.16$).

\begin{figure}[!htbp]
\centering
\includegraphics[width=\textwidth]{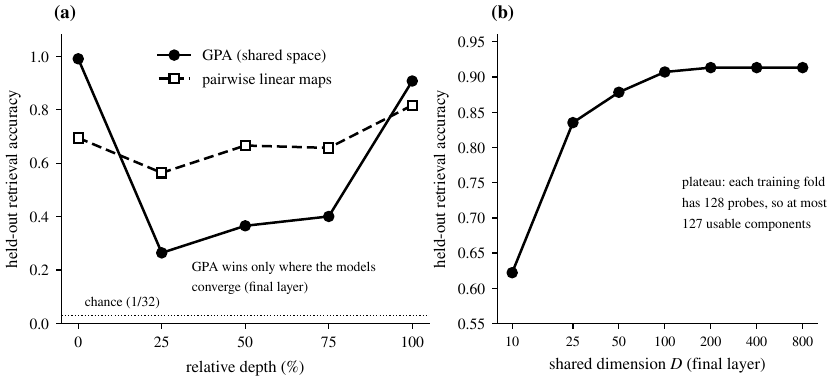}
\caption{(a) Mean held-out retrieval accuracy by depth, set 1, for GPA at $D=100$ and for pairwise linear maps. GPA wins only at the final layer and at the (lexical) embedding layer. (b) Accuracy against the shared dimension $D$ at the final layer. The plateau coincides with the rank limit of the 128-probe training folds.}
\label{fig:gpa}
\end{figure}

\paragraph{GPA is not better at every depth.} Figure~\ref{fig:gpa}a shows that in the middle of the network the pairwise maps are far better ($0.56$ to $0.67$ against $0.26$ to $0.40$ at $25$--$75\%$ depth), in line with the earlier finding that the middle layers of these models are individual. A rotation inside the top $D$ principal directions is a much smaller hypothesis class than an unconstrained linear map. It pays off only where the models have converged, and in this study that is the final layer. The advantage at the embedding layer ($0.99$) is the lexical-overlap artifact discussed above. For these reasons we construct the shared space at the final layer only.

\paragraph{Dimension.} Accuracy grows with $D$ and is flat from $D=200$ upward (Figure~\ref{fig:gpa}b). We initially read this as a property of the models. It is not established: each training fold has 128 probes, so centered PCA has at most 127 informative components, and any larger $D$ adds only zero-variance directions. The plateau reflects our sample size. We still use $D=200$ for the released space because it is fit on all 400 probes.

\paragraph{Released artifact.} The released module fits once on all 400 probes (both sets, all ten categories) and exposes a function that maps a final-layer activation of any of the five models to a point in the 200-dimensional space.

\section{Stage 3: Where Does the Shared Space Stop Working?}
\label{sec:stage3}

The space of the previous section was first fit on the eight prose categories. We asked how far it generalizes. In the content-type tests below, accuracy is retrieval accuracy within a pool of 10 probes of one type (chance $0.10$).

\paragraph{Unseen categories.} Code and multilingual probes, never used in fitting, reached $0.29$ to $0.36$ accuracy in pairwise retrieval (roughly $6$ to $14$ times chance in that test), far below the in-distribution $0.9$ range.

\begin{figure}[!htbp]
\centering
\includegraphics[width=\textwidth]{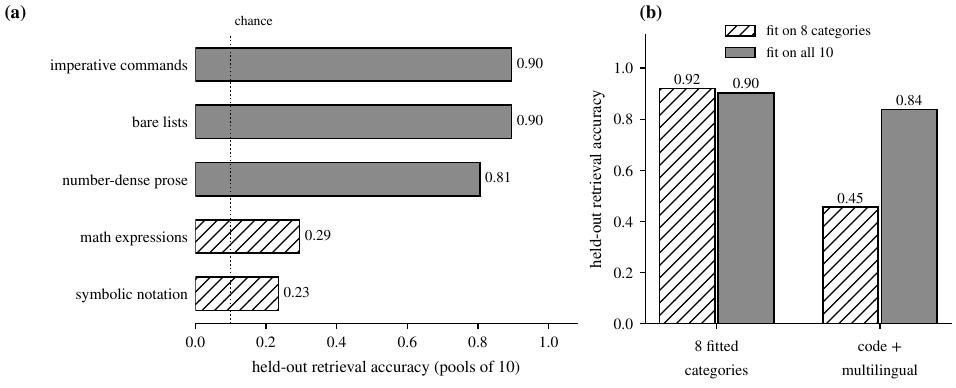}
\caption{(a) Held-out retrieval accuracy of a space fit on eight categories, for content types it never saw (pools of 10, chance $0.10$). Hatched bars are the failures. (b) Matched five-fold comparison with and without code and multilingual probes in the fit.}
\label{fig:boundary}
\end{figure}

\paragraph{Content unlike any fitted category.} Ten each of imperative commands, bare enumerated lists and mathematical expressions were written for this test, none resembling a fitted category. Imperatives and lists reached $0.90$ (Figure~\ref{fig:boundary}a). Mathematical expressions reached only $0.30$. A follow-up separated numerals from symbols: number-dense ordinary prose scored $0.81$ and symbol-dense notation with fewer numerals scored $0.24$. The boundary is therefore not ``resembles a fitted category'' but closer to ``ordinary natural-language vocabulary versus non-standard symbolic notation''. We stress that this is a characterization of a few content types with ten items per type, not a proof of where the boundary lies.

\paragraph{Refitting.} In a matched five-fold comparison, adding code and multilingual probes to the fit nearly doubled their held-out accuracy ($0.45\rightarrow0.84$) at a cost of $0.02$ on the eight original categories ($0.92\rightarrow0.90$; Figure~\ref{fig:boundary}b). The released space was refit on all ten categories.

\section{Stage 4: From Representations to Task Performance}
\label{sec:stage4}

The previous sections concern representations only. We now ask whether the space is useful for choosing among the models. All five models were run with a few-shot prompt and greedy decoding (they are base models) on questions with unambiguous answers, and outputs were graded by exact match.

\subsection{A signal that was a confound}

\begin{figure}[!htbp]
\centering
\includegraphics[width=\textwidth]{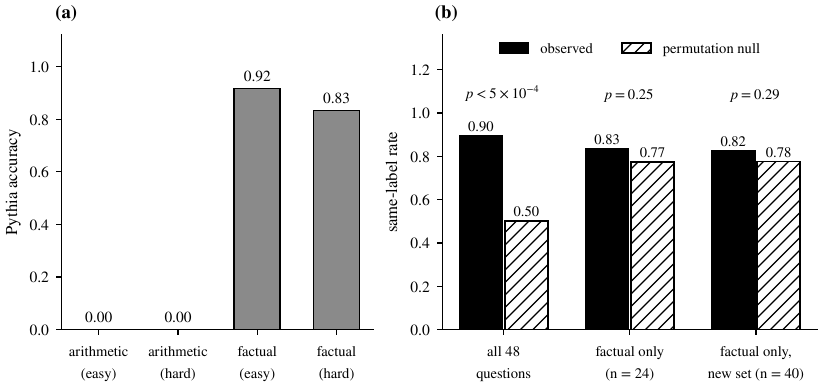}
\caption{(a) Pythia's accuracy by question category on the 48-question set (12 per cell). (b) Nearest-neighbor same-correctness-label rate in the shared space against the permutation-null mean. The strong result on all questions disappears when only factual questions are used.}
\label{fig:confound}
\end{figure}

For Pythia, a question's position in the shared space appeared to predict whether Pythia answered it correctly. The nearest-neighbor same-label rate was $0.896$ against a null of $0.500$ ($p<5\times10^{-4}$) on a 48-question set, and a similar result held on an earlier 24-question set. A direct check removed it. Pythia's correctness followed question category almost exactly: $0/24$ correct on arithmetic (easy and hard alike) and $21/24$ on factual questions (Figure~\ref{fig:confound}a). Because the space is already known to separate categories, ``predicts correctness'' reduces to ``predicts category''. Restricted to factual questions, where Pythia's answers varied, the same-label rate was $0.833$ against a null of $0.773$ ($p=0.25$, $n=24$), and on a second, independently written and harder factual set it was $0.825$ against $0.776$ ($p=0.29$, $n=40$; Figure~\ref{fig:confound}b). We found no fine-grained correctness signal. Both factual-only tests have few incorrect answers ($3$ and $5$), so they have low power, and we do not claim the signal is absent, only that we could not detect it.

\subsection{A router, and four searches for a case where it helps}

We built the simplest router that the confirmed signal supports. A nearest-centroid classifier predicts a question's category (arithmetic or factual) from its position in the shared space, and the router sends the question to the model with the highest accuracy for that category on training data. Trained on one question set and tested on another, category classification reached $0.979$ accuracy, but the routing decision became a single model for every category (Qwen, tied with SmolLM2), the same as always using the best model.

We then searched four task types for any case where a different model should be preferred (Table~\ref{tab:tasks}).

\begin{table}[!htbp]
\centering
\caption{Accuracy of each model on four task designs. In no row does a model other than Qwen or SmolLM2 come out on top.}
\label{tab:tasks}
\begin{tabular}{lrrrrr}
\toprule
Task type ($n$ questions) & Qwen & Pythia & Mamba & SmolLM2 & OLMo \\
\midrule
Simple QA (24) & 100\% & 46\% & 92\% & 100\% & 0\% \\
Mixed-difficulty QA (48) & 98\% & 44\% & 75\% & 100\% & 0\% \\
Multi-step reasoning (30) & 77\% & 37\% & 40\% & 73\% & 0\% \\
Associative recall (24) & 100\% & 54\% & 46\% & 96\% & 0\% \\
\bottomrule
\end{tabular}
\end{table}

\begin{figure}[!htbp]
\centering
\includegraphics[scale=0.88]{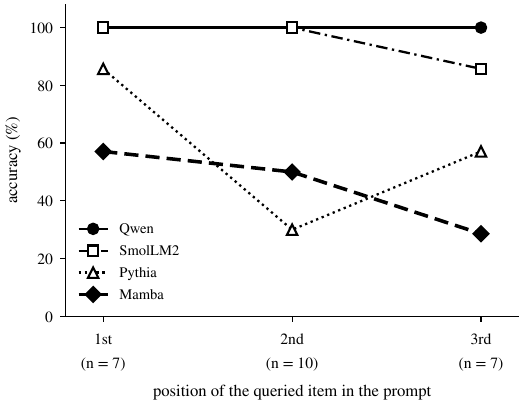}
\caption{Associative-recall accuracy by which of three mentioned items was queried. OLMo (0\% everywhere) is omitted. With $n=7$ to $10$ per position, differences within a model are not statistically distinguishable; Mamba's decline from first to last is suggestive only.}
\label{fig:recall}
\end{figure}

The fourth task was chosen deliberately. Given the documented weakness of state-space models at associative recall \citep{gu2023mamba,arora2023zoology}, we expected Mamba to do worse than the transformers, and it did: this is the only task where it fell below Pythia. We tagged each of the 24 key-value questions by whether the queried item was the first, second or third mentioned ($n=7,10,7$). Mamba was best on the first item ($57.1\%$) and worst on the last ($28.6\%$), while Qwen and SmolLM2 were at or near ceiling in every position and Pythia was erratic (Figure~\ref{fig:recall}). We had expected the opposite direction, a recency advantage, since the latest input should be freshest in a compressed state. We report what we saw. With seven items per cell, the difference between $4/7$ and $2/7$ is well within chance variation, so this primacy pattern is a hypothesis for replication, not a finding.

OLMo scored $0\%$ on every task. Before treating this as real we checked next-token logits: this checkpoint's greedy completions barely varied with question content and gave an unrelated generic string. A longer few-shot prompt and plain completion-style prompts without question-answer framing gave the same result, which rules out prompt formatting. We treat it as a checkpoint-specific limitation under greedy decoding, not as a statement about OLMo-family models.

\subsection{Closing the routing question}

Across four task designs (factual recall, arithmetic, compositional reasoning, associative lookup) we found no case in which one of the three weaker models beats the stronger generalists. For this model set and these tasks, routing has no demonstrated value over always using the best single model. We call this a negative result and not an underpowered search, because the final task produced an explanatory result and the four tasks differ substantially in demand. We cannot rule out a crossover on other domains, such as code generation, multilingual tasks or long-context tasks.

\section{Discussion and Limitations}
\label{sec:discussion}

The central positive result is that models with different architectures, trained by different groups, organize the same input in measurably similar ways at their final layer, including a state-space model that shares no sequence-mixing mechanism with the rest, and that this correspondence is strong enough to support a usable shared space from orthogonal maps alone. It is consistent with the convergence argument of \citet{huh2024platonic}, though we test only five small models and cannot speak to scaling.

We caution against two readings. First, Mamba's lower representational agreement and its weaker associative recall are separate observations from separate experiments. We did not test any causal link between them, and we do not claim that one explains the other. Second, the shared space was fit without any task labels and, as Stage~4 shows, it carries category information but no detectable information about whether a given model gets a given question right.

Limitations are as follows.
\begin{itemize}
  \item \textbf{Scale.} All models are $1$--$2$B-parameter base checkpoints. Instruction-tuned or larger models were not tested. Probe sets have 200 inputs and most task sets have 24 to 48 questions, so several nulls (notably the within-category correctness test and the position-wise recall breakdown) are low-powered.
  \item \textbf{Authored probes.} Both probe sets were written by the authors. A set drawn from natural corpora would remove the possibility that our writing style makes the models agree more than they otherwise would. The independent second set reduces but does not remove this concern.
  \item \textbf{Retrieval pools.} Retrieval accuracy is measured within pools of $10$ or $32$ items, so absolute numbers are not comparable to retrieval over large databases.
  \item \textbf{Dimension.} The plateau in $D$ is a rank limit set by our sample size, so we do not know how many dimensions carry cross-model structure.
  \item \textbf{Pooling.} Uniform mean pooling may understate Mamba.
  \item \textbf{Single run.} Every experiment used one pass of one model checkpoint with greedy decoding. We did not vary seeds beyond cross-validation fold assignment.
  \item \textbf{Task coverage.} The negative routing result is specific to these five models and four task types.
\end{itemize}

\section{Conclusion}

Five frozen language models from four organizations, including a state-space model, share real and replicable representational structure that can be put into one space by orthogonal rotations alone, with no model privileged. At the final layer the space recovers $0.91$--$0.93$ cross-model retrieval accuracy and generalizes to ordinary English of unfamiliar grammatical form, but not to mathematical or logical notation, and not in the middle layers. It does not, for this model set and these tasks, provide a better way to pick which model to use. We report both outcomes, including two apparent signals (category-level agreement and correctness prediction) that did not survive controls.

\paragraph{AI assistance.} The experiments, analysis and drafting were carried out with substantial assistance from Claude (Anthropic), an AI model, which helped write code, run experiments, check statistics and draft text. The human authors designed the study, directed each step, reviewed all results and are responsible for the content.


\bibliographystyle{plainnat}
\bibliography{references}

\end{document}